\documentclass[journal]{IEEEtran}

\usepackage[utf8]{inputenc} 
\usepackage[T1]{fontenc}    
\usepackage{hyperref}       
\usepackage{url}            
\usepackage{booktabs}       
\usepackage{amsfonts}       
\usepackage{nicefrac}       
\usepackage{microtype}      
\usepackage{graphicx}

\usepackage[backend=bibtex]{biblatex}
\bibliography{smallbiblio}
\AtEveryBibitem{%
  \clearlist{language}%
}

\title{Neural networks with differentiable structure}

\author{
Thomas Miconi\\
  The Neurosciences Institute\\
 La Jolla, CA, USA \\
  \texttt{miconi@nsi.edu} \\
}

\begin{document}
 
\maketitle

\begin{abstract}

While gradient descent has proven highly successful in learning connection
weights for neural networks, the actual structure of these networks is usually determined by hand, or by
other optimization algorithms.  Here we describe a simple method to make
network structure differentiable, and therefore accessible to gradient descent.
We test this method on recurrent neural networks applied to simple
sequence prediction problems. Starting with initial networks containing only
one node, the method automatically grows networks that successfully solve the
tasks. The number of nodes in the final network correlates with task
difficulty. The method can dynamically increase network size in response to an
abrupt complexification in the task.
Variable-size networks grown with the method outperform fixed-size
networks of higher, lower or identical size, hinting at a possible advantage of growing networks. We conclude by discussing how this
method could be applied to more complex networks, such as feedforward layered
networks, or multiple-area networks of arbitrary shape.

\end{abstract}

\section{Introduction}

Neural networks are usually optimized by applying some form gradient descent to
the numerical parameters of a fixed connectivity graph. This method can
successfully train very large networks for complex tasks. However, the actual
structure of the network itself (number of neurons, connectivity graph, etc.) is usually not modified by the gradient
descent algorithm. Most often, network structure is designed by hand, in a delicate process of parameter tuning.
When network structure is optimized, it is generally with a different
algorithm, including evolutionary techniques such as NEAT \cite{Stanley2002-ug} or heuristic-based methods such as HyperOpt \cite{Yamins2014-us}.

Manual design of network structure is time-consuming and subject to arbitrary
choices that may or may not reflect the demands of the task at hand.
Furthermore, letting the size of the network grow autonomously may actually
improve learning performance, as posited in the NEAT framework
\cite{Stanley2002-ug}.  It would therefore be desirable to extend the process
of gradient descent to network structure itself. This requires making network
structure differentiable, at least to a usable approximation. Here we describe
a simple method for performing gradient descent over network structure, and
show that this method can adaptively design recurrent networks of a few dozen
units for simple sequence prediction tasks.

\section{Method}

\subsection{Description of the algorithm}

Here we describe our method, in the context of recurrent networks with
all-to-all potential connectivity (in the conclusion, we suggest how the method
could be extended to more complex architectures, including layered feedforward
networks). In this situation, structure is determined by the number of nodes in
the network $N$, which automatically determines the connectivity graph as a
simple square matrix of size $N*N$. Our goal is to make the number of nodes
differentiable and amenable to gradient descent and backpropagation. 

The first step in our method is to impose a penalty on the L1-norm (sum of absolute values) of \textit{outgoing} weights from each neuron. This includes both lateral and feedforward weights.
As is well-known, minimizing the L1-norm 
tends to concentrate the remaining total weight among the fewest possible
elements, in comparison to Euclidean L2-norm minimization.  As a result,
backpropagation will tend to minimize the number of neurons with non-zero total output, and
thus of ``active'' neurons: each neuron must ``earn its keep'', by contributing
to overall network performance, to counter-balance the effect of L1-norm
minimization, or else face effective ``soft'' deletion by having its outgoing weights 
fall to zero.\footnote{Importantly, note that L1 regularization on outgoing weights is quite different from directly imposing an
    L1 regularization on neuron activities themselves. L1 regularization of
    neuron activities ensures that few neurons will be active \textit{at any
    given time}, but does not ensure that any neuron will become fully silent
    over extended time.  Instead, L1 regularization of neuron activities may
    encourage neurons to distribute and decorrelate their activations other
    time so that each neuron responds to a small proportion of inputs; this is
    precisely the (intended) effect of L1-regularization in \textit{sparse
coding} schemes \cite{Olshausen1996-vz}. By contrast, penalizing outgoing weights can truly
turn neurons ``on'' or ``off'' in a time-independent fashion: a neuron with 
zero output weights is guaranteed to be silent for any input. }

This method creates a ``soft''
structural variability, whereby gradient descent tries to solve the task at
hand under the constraint of minimizing the number of neurons with non-zero
outgoing weights. We want to turn these ``soft'' structure changes into hard
structural changes in the actual number of neurons and size of the weight
matrix. To this end, we first specify a \textit{deletion threshold} $T_D$, such
that any neuron for which the L1-norm of outgoing weights falls below this threshold is marked for
potential deletion. Then, we simply specify that at any given time, the network
must only contain a fixed, small number $k$ of neurons below the deletion
threshold. If the number of sub-threshold neurons exceeds $k$, then ``excess''
sub-threshold neurons are actually deleted from the network.  Conversely, if
backpropagation finds it necessary to inflate neuron output weights to the extent that
fewer than $k$ neurons have sub-threshold output weight norm, then we add a new neuron
to the simulation, with initially random connectivity and outgoing weights
initially chosen to have L1-norm exactly equal to the deletion threshold. Note that, because the threshold
value is low, new neurons initially have a very small effect on overall network
behavior.

This mechanism allows backpropagation to adjust network size to problem
demands. If more neurons are needed to solve the problem at hand,
backpropagation will simply expand the outgoing weights of currently sub-threshold
neurons, so as to allow them to have an impact on output computation, while adjusting their connectivity. By
contrast, if new neurons fail to contribute to network performance,
L1-minimization  will reduce their outgoing weights and eventually drive them below
deletion threshold. The sub-threshold neurons thus act as a computational
reserve, ready to be mobilized if the problem at hand demands it.

Finally, as a stabilization measure, we make
addition and deletion probabilistic, so that whenever a neuron is to be added
or deleted, the event only occur with a certain fixed probability $P_{add}$ or
$P_{del}$. As a result, the network will occasionally possess more or less than
$k$ subthreshold neurons. All networks in our experiment start with only one
node, following the philosophy of ``augmenting topologies'' expounded in NEAT
\cite{Stanley2002-ug}.

\subsection{Implementation details}

Our implementation is based on Andrej Karpathy's \verb+min-char-rnn.py+ and
inherits most of its parameters.  The networks are trained for 100000 cycles,
where each cycle consists of reading a sequence of 40 characters while trying
to predict the next character, followed by a parameter update based on
backpropagation through time. Network output is provided by a single output
layer with 4 nodes (one per possible character), each of which reports the
predicted probability that the corresponding character is next in the sequence. The output layer is fully connected with the variable-size recurrent layer.
Loss is defined as cross-entropy between the predicted distribution and the
actual (one-hot) outcome.  Any addition or deletion also occurs at the same time as parameter
update (that is, at the end of each successive 40-char sequence). 


There are thus 5 additional parameters in our method: $k$, $T_D$,  $P_{add}$,
$P_{del}$, and $A_{L1reg}$ (the strength of the L1-norm penalty 
over the weights). In all simulations shown here, those were set to $k=1$,
$T_D=0.05$,  $P_{add}=0.01$, $P_{del}=0.05$, and
$A_{L1reg}=10^{-4}$.

All code is available on GitHub at \url{https://github.com/ThomasMiconi/DiffRNN}.

\section{Experiments}

\subsection{Tasks}

To test the plausibility of our method, we choose two simple sequence prediction
problems. In each problem, the task of the network is to predict the next
character in an ongoing sequence of characters. Both problems use the same
alphabet, consisting of characters $a$, $b$, $($ and $)$. 

The first problem (``easy problem'') is composed of groups of one or more $ab$
digraphs, enclosed in matching parentheses. After every $ab$ digraph, there is
a constant probability of adding an additional $ab$ digraph (p=0.75), or to close the
group with a closing parenthesis instead (p=0.25). Thus the number of digraphs in each
group follows an exponential distribution. A typical sequence looks like this:

\begin{center}
$(abab)(ab)(ab)(ababab)(abababababab)(abab)(abababab)\ldots$
\end{center}

Note that the problem is highly constrained: the only choice occurs after a
$b$, when the network must decide whether to insert a $)$ or an $a$, which has
a well-defined probability. Every other choice is unambiguously specified by the problem.

The second problem (''hard problem'') is composed of groups of six letters
enclosed in matching parentheses. The rule is that each new group must be the
reverse of the previous group, with one randomly chosen letter changed. A
typical sequence looks like this:

\begin{center}
$(aabbab)(babaaa)(aaabbb)(bbaaaa)(abaabb)(baaaba) \ldots$
\end{center}

To reach optimal performance on this task, the network must maintain a memory
of the previous sequence of six characters, and then reverse it, in addition to opening and closing parentheses. This is a
 more difficult problem than the previous one, and thus we expect that
optimal networks for either task would look quite different from each other.

\section{Results}

\begin{figure}[ht]
\label{fig:easyandhard}
  \centering
\includegraphics[scale=0.9]{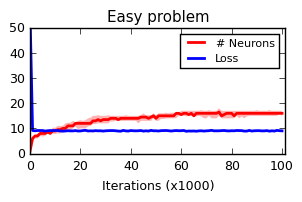}
\includegraphics[scale=0.9]{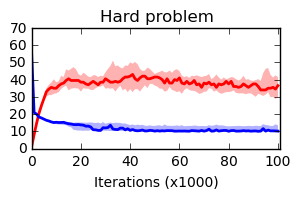}
  \caption{Model performance on an easy task (left panel) and a hard task
(right panel). Both performance (cross-entropy loss between predicted and
actual character) and number of neurons are shown as a function of time. Dark curves and shaded areas indicate median and inter-quartile range over 20 runs, respectively. The
model settles on larger network size for the more complex problem. }
\end{figure}

\subsection{Performance and network size in hard and easy tasks}

Results are shown in Figure \ref{fig:easyandhard}. We show both median
performance (cross-entropy loss) and median number of neurons as a function of
time, over 20 runs. As expected, the hard problem leads to somewhat higher loss
than the easy problem. Importantly, the hard problem elicits larger networks
than the easy problem (37 neurons vs. 14 neurons after 100000 learning cycles).
Thus, the algorithm appropriately allocated more neurons to solve a more difficult
task.

\begin{figure}[t]
\label{fig:fixedsize}
  \centering
\includegraphics[scale=0.9]{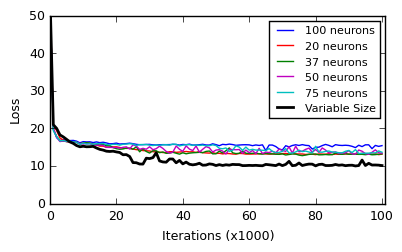}
  \caption{Comparison of performance for variable and fixed size, for the
``hard'' problem. The thick black line shows variable-size network performance
and is identical to the blue curve in Fig. \ref{fig:easyandhard}, right panel.
Thin colored curves indicate performance of fixed-size networks of various
sizes. Curves show medians over 20
runs; inter-quartile ranges (not shown for clarity) are comparable to those
seen in Fig. \ref{fig:easyandhard}. Variable-size networks outperform fixed-size
networks for the problem described here. }
\end{figure}

An important question is whether the use of variable-size networks has an
impact on performance. We compared the performance of our algorithm against
fixed-size networks with various numbers of neurons, ranging from 10 to 100,
including one with the same network size as was eventually preferred by our
algorithm (i.e. 37 neurons).  Results are shown in figure \ref{fig:fixedsize},
again showing the median loss among 20 runs as a function of time.
Intriguingly, the variable-size network actually outperforms fixed-size
networks of any size. This result may reflect the advantages of  ``augmenting
topologies'' (starting with a minimal network and only adding complexity as
needed), as expounded in NEAT \cite{Stanley2002-ug}, at least for the simple
problems tackled here.

\subsection{Dynamical adjustment of network size in response to changing conditions}

\begin{figure}[b]
\label{fig:easyhardeasy}
  \centering
\includegraphics[scale=0.9]{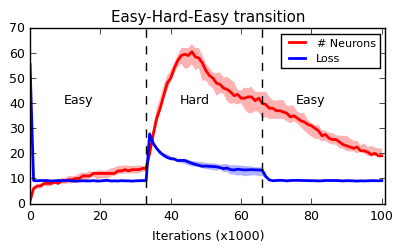}
  \caption{Dynamic adjustment of network size in response to abrupt complexification and simplification of an ongoing task.}
\end{figure}

What happens if task difficulty suddenly changes? We tested our network by
 switching from the ``easy'' to the ``hard'' sequence after 33000
cycles, and then back again to the ``easy'' sequence after 66000 cycles.
Results are shown in Figure \ref{fig:easyhardeasy}. Interestingly, the network
successfully handles the abrupt complexification of the problem by allocating
more neurons. Following a large increase, the network then sheds off excess
neurons, without damaging performance. This process continues when the problem
switches back to the ``easy'' sequence (note that performance quickly returns
to optimal levels). Thus, the network successfully adapts its size to the complexity of the problem at hand.

\section{Conclusions and future work}

We have described a method through which the size of a recurrent network can be
modified by gradient descent. The method described here can successfully build
networks of appropriate size to handle simple problems. This simple method
immediately suggests several alternatives and possible extensions.

For example, deletion of neurons could be biased by neuron ``age'' (i.e. how
long the neuron has been present), rather than being random. Deleted neurons
could be partially preserved, so that newly added neurons could actually
inherit connectivity of previously deleted ones, rather than being randomly
initialized. Such adaptations were not necessary for the problems considered
here, but might be considered in future applications to more challenging tasks.

The method described here extends naturally to layered feedforward networks.
Within each layer, the method can be applied essentially unchanged to adjust
layer size. The number of layers can also be made differentiable, by adding and deleting
\textit{residual} layers \cite{He2015-gk} with initially low 
pre-additive output weights. These residual layers, which would initially have minimal impact
on the network's output, would play the same role as sub-threshold neurons in
the method described above.
 Similarly, by considering each layer as a higher-order
``node'', subject to a global outgoing norm penalty, the method described above could in
principle be extended to arbitrary networks composed of multiple areas, with
arbitrary connectivity between areas. Further work is needed to assess the
practicality of these and other possible extensions.

\small

\printbibliography

\end{document}